\documentclass{article}

\usepackage{times}
\usepackage{graphicx} 
\usepackage{booktabs}       
\usepackage{subfigure} 
\usepackage{multirow}

\usepackage{natbib}

\usepackage{algorithm}
\usepackage{algorithmic}
\usepackage{hyperref}
\usepackage{url}
\usepackage{graphicx}
\usepackage{float}
\usepackage{amsmath, amsfonts}
\usepackage{amsthm}
\usepackage{varwidth}
\usepackage{calc} 
\newtheorem{theorem}{Theorem}[section]
  	
\usepackage[accepted]{icml2017}

\makeatletter
\g@addto@macro\normalsize{%
  \abovedisplayskip 5pt plus 2pt minus 4pt%
  \belowdisplayskip \abovedisplayskip
  \abovedisplayshortskip 4pt plus3pt%
  \belowdisplayshortskip 4pt plus3pt minus3pt%
}
\makeatother

\icmltitlerunning{Maximum-Likelihood Augmented Discrete Generative Adversarial Networks}

\begin{document} 

\twocolumn[
\icmltitle{Maximum-Likelihood Augmented Discrete Generative Adversarial Networks}



\icmlsetsymbol{equal}{*}

\begin{icmlauthorlist}
\icmlauthor{Tong Che}{equal,udm}
\icmlauthor{Yanran Li}{equal,polyu}
\icmlauthor{Ruixiang Zhang}{equal,hkust}
\icmlauthor{R Devon Hjelm}{udm,ivado}
\icmlauthor{Wenjie Li}{polyu} 
\icmlauthor{Yangqiu Song}{hkust}
\icmlauthor{Yoshua Bengio}{udm}
\end{icmlauthorlist}

\icmlaffiliation{udm}{Montreal Institute for Learning Algorithms, Universit\'{e} de Montr\'{e}al, Montr\'{e}al, QC H3T 1J4, Canada}
\icmlaffiliation{polyu}{Department of Computing, The Hong Kong Polytechnic University, Hong Kong}
\icmlaffiliation{hkust}{The Hong Kong University of Science and Technology}
\icmlaffiliation{ivado}{IVADO}

\icmlcorrespondingauthor{Tong Che}{tong.che@umontreal.ca}

\icmlkeywords{generative adversarial networks, text generation, reinforcement learning}

\vskip 0.3in
]



\printAffiliationsAndNotice{\icmlEqualContribution} 

\begin{abstract} 
Despite the successes in capturing continuous distributions, the application of generative adversarial networks (GANs) to discrete settings, like natural language tasks, is rather restricted. The fundamental reason is the difficulty of back-propagation through discrete random variables combined with the inherent instability of the GAN training objective. To address these problems, we propose Maximum-Likelihood Augmented Discrete Generative Adversarial Networks. Instead of directly optimizing the GAN objective, we derive a novel and low-variance objective using the discriminator's output that follows corresponds to the log-likelihood. Compared with the original, the new objective is proved to be consistent in theory and beneficial in practice. The experimental results on various discrete datasets demonstrate the effectiveness of the proposed approach.
\end{abstract} 

\section{Introduction}

Generative models are appealing because they provide ways to obtain insights on the underlying data distribution and statistics. In particular, these models play a pivot role in many natural language processing tasks such as language modeling, machine translation, and dialogue generation. However, the generated sentences are often unsatisfactory~\citep{Sordoni2015ANN,bowman2015generating,serban2016vhred,wiseman2016sequence}. For example, they often lack of consistency in long-term semantics and have less coherence in high-level topics and syntactics~\citep{bowman2015generating,Zhang2016VariationalNM}.

This is largely attributed to the defect in the dominant training approach for existing discrete generative models. To generate discrete sequences, it is popular to adopt auto-regressive models through teacher forcing~\citep{teacherforcing} which, nevertheless, causes the \textit{exposure bias} problem~\citep{ranzato2015sequence}. The existing approach trains auto-regressive models to maximize the conditional probabilities of next tokens based on the ground-truth histories. In other words, during training, auto-regressive generative models are only exposed to the ground truths from the data distribution rather than those from the model distribution, i.e., its own predictions. It prohibits the trained model to take advantage of learning in the the context of its previous generated words to make the next prediction, resulting in a bias and difficulty in approaching the true underlying distribution~\citep{ranzato2015sequence,bengio2015scheduled}. Another limitation of teacher forcing is that it is inapplicable to those auto-regressive models with latent random variables, which have performed better than autoregressive (deterministic state) recurrent neural networks (i.e. usual RNNs, LSTMs or GRUs) on multiple tasks~\cite{serban2016vhred,Miao2016NeuralVI,Zhang2016VariationalNM}.

An alternative and attractive solution to training autoregressive models is using generative adversarial networks (GAN)~\citep{gan}. The above discussed problem can be prevented if the generative models were able to visit its own predictions during training and had an overall view on the generated sequences. We suggest to facilitate the training of autoregressive models with an additional discriminator under the GAN setting. With a discriminator trained to separate real versus generated sequences, the generative model is able to make use of the knowledge of the discriminator to improve itself. Since the discriminator is trained on the entire sequence, it can in principle provide the training signal to avoid the problem of exposure bias.  

However, it is nontrivial to apply GANs to discrete data as it is difficult to optimize of the generator using the signal provided by the discriminator. In fact, it is usually very hard to push the generated distribution to the real data distribution, if not impossible, by moving the generated sequence (e.g., a faulty sentence) towards a ``true'' one (e.g., a correct sentence) in a high dimensional discrete state space. As standard back-propagation fails in discrete settings, the generator can be optimized using the discriminator's output as a reward via reinforcement learning. Unfortunately, even with careful pre-training, we found that the policy has difficulties to get positive and stable reward signals from the discriminator.  

To tackle these limitations, we propose \textbf{Ma}ximum-\textbf{Li}kelihood Augmented Discrete \textbf{G}enerative \textbf{A}dversarial \textbf{N}etworks (\textbf{MaliGAN}). At the core of this model is the novel GAN training objective which sidesteps the stability issue happening when using the discriminator output as a direct reinforcement learning reward. Alternatively, we develop a normalized maximum likelihood optimization target inspired by~\cite{rewardaug}. We use importance sampling and several variance reduction techniques in order to successfully optimize this objective. The procedure was discovered independently
from us by~\citet{hjelm2017boundary} in the context
of image generation.

The new target brings several attractive properties in the proposed MaliGAN. First, it is theoretically consistent and easier to optimize (Section~\ref{sec:analysis}). Second, it allows the model not only to maximize the likelihood of good behaviors, but also to minimize the likelihood of bad behaviors, with the help of a GAN discriminator.  Equipped with these strengths, the model focuses more on improving itself by gaining beneficial knowledge that is not yet well acquired, and excluding the most probable and harmful behaviors. Combined with several proposed variance reduction techniques, the proposed MaliGAN successfully and stably models discrete data sequences (Section~\ref{sec:exp}). 

\section{Preliminaries and Overview}
The basic framework for discrete sequence generation is to fit a set of data $\{\mathbf{x}_i\}_{i=1}^N$ coming from an underlying generating distribution $p_d$ by training a parameterized auto-regressive probabilistic model $p_\theta$. 

In this work, we aim to generate discrete data, especially discrete sequential data, under the GAN setting~\citep{gan}. GAN defines a framework for training generative models by posing it as a minimax game against a discriminative model. The goal of the generator $G$ is to match its distribution $p_g$ to the real data distribution $p_{d}$. To achieve this, the generator transforms noise $z$ sampled from $p(z)$ to a data sample $G(z)$. Following this, the discriminator $D$ is trained to distinguish between the samples coming from $p_{d}$ and $p_g$, and can be used to provide a training signal to the generator.

When applying the GAN framework to discrete data, the discontinuity prohibits the update of the generator parameters via standard back-propagation. To tackle this, one way is to employ a typical reinforcement learning (RL) strategy that directly uses the GAN discriminator's output, $D$ or $\log D$ as a reward. In practice, the problem is usually solved by REINFORCE-like algorithms~\cite{reinforce}, perhaps with some variance reduction techniques. 

Formally, we train a generator $G(\mathbf{x})$ together with a discriminator $D(\mathbf{x})$. In its original form, the discriminator is trained to distinguish between the generating distribution $p_\theta$ and the real data distribution $p_d$. The generator is then trained to maximize $\mathbb{E}_{\mathbf{x}\sim p_\theta}[\log D(\mathbf{x})]$. Namely, the objective for the generator to optimize is as follows:

\small
\begin{equation*}
\begin{aligned}
\mathcal{L}_{GAN}(\theta) &= -\mathbb{E}_{\mathbf{x}\sim p_\theta}[\log D(\mathbf{x})] \\
&\approx  -\frac{1}{n} \sum_{i=1}^n \log D(\mathbf{x}_i), \;\;\; \mathbf{x}_i \sim p_\theta.
\end{aligned}
\end{equation*}
\normalsize

Our work is related to the viewpoint of casting the GAN training as a reinforcement learning problem with a moving reward signal monotone in $D(\mathbf{x})$.
Define the normalized probability distribution $q'(\mathbf{x}) = \frac{1}{Z(D)}D(\mathbf{x})^{1/\tau}$ in
some bounded region to guarantee integrability (note that $D$ is an approximation to $\frac{p_d}{p+p_d}$ if $D$ is well trained) and also put a maximum-entropy regularizer $\mathbb{H}(p_\theta)$ to encourage diversity, yielding the regularized loss:

\small
\begin{equation}
\begin{aligned}
\mathcal{L}_{GAN}(\theta) &= -\mathbb{E}_{\mathbf{x}\sim p_\theta}[\log D(\mathbf{x})] -\tau \mathbb{H}(p_\theta)\\
&=\tau \text{KL}(p_\theta||q') + c(D)
\end{aligned}
\label{eq:3}
\end{equation}
\normalsize

where \emph{c(D)} is a constant depending only on $D$. Hence, optimizing the traditional GAN is basically equivalent to optimizing the KL-divergence $\text{KL}(p_\theta||q')$. One major problem with this approach is that $q'$ always moves with $D$, which is undesirable for both stability and convergence. When we have some samples $\mathbf{x}_i\sim p_\theta$, we want to change $\theta$ a bit in order to adjust the likelihood of samples $\mathbf{x}_i$ to improve the quality of the generator. However, since initially $p$ generates very bad sequences, it have little chance of generating good sequences in order to get positive rewards. Though the dedicated pre-training and variance reduction mechanisms help~\citep{seqgan}, the RL algorithm based on the moving reward signal still seems very unstable and does not work on large scale datasets.

We therefore propose to utilize the information of the discriminator as an additional source of training signals, on top of the maximum-likelihood objective. We employ importance sampling to make the objective trainable. The novel training objective has much less variance than that in vanilla reinforcement learning approaches that directly adopt $D$ or $\log D$ as reward signals. The analysis and discussions will be presented in more detail in Section~\ref{sec:analysis}.

\section{Maximum-Likelihood Augmented Discrete Generative Adversarial Networks}

In this section, we present the details of the proposed model. At the heart of this model is a novel training objective that significantly reduces the variance during training, including the theoretical and practical analysis on the objective's equivalence and attractive properties. We also show how this core algorithm can be combined with several variance reduction techniques to form the full MaliGAN algorithm for discrete sequence generation.  

\subsection{Basic Model of MaliGAN}
We propose Maximum-Likelihood Augmented Discrete Generative Adversarial Networks (MaliGAN) to generate the discrete data. With MaliGAN, we train a discriminator $D(\mathbf{x})$ with the standard objective that GAN employs. What is different from GANs is a novel objective for the generator to optimize, using importance sampling, which makes the training procedure closer to maximum likelihood (MLE) training of auto-regressive models, and thus being more stable and with less variance in the gradients.

To do so, we keep a delayed copy $p'(\mathbf{x})$ of the generator whose parameters are updated less often in order to stabilize training. From the basic property of GANs, we know that an optimal $D$ has the property $D(\mathbf{x})=\frac{p_d}{p_d+p'}$. So in this case, we have $p_d = \frac{D}{1-D}p'$. Therefore, we set the target distribution $q$ for maximum likelihood training to be $\frac{D}{1-D}p'$. Let $r_D(\mathbf{x}) = \frac{D(\mathbf{x})}{1-D(\mathbf{x})}$, we define the augmented target distribution as: 
\begin{align*}
\small
q(\mathbf{x})=\frac{1}{Z(\theta')}\frac{D(\mathbf{x})}{1-D(\mathbf{x})}p'(\mathbf{x})=\frac{r_D(\mathbf{x})}{Z(\theta')}p'(\mathbf{x}) 
\end{align*}

Regarding $q$ as a fixed probability distribution, then the target to optimize is:
\begin{align*}
\small
L_{G}(\theta) = \text{KL}(q(\mathbf{x})||p_\theta(\mathbf{x}))
\end{align*}

This objective has an attractive property that $q$ is a ``fixed'' distribution during training, i.e., if $D$ is sufficiently trained, then $q$ is always approximately the data generating distribution $p_d$. By defining the gradient as $\nabla L_{G} = \mathbb{E}_q[\nabla_\theta\log p_\theta (\mathbf{x})]$, we have the following importance sampling formula:
\begin{align*}
\small
\nabla L_{G} &= \mathbb{E}_{p'}[\frac{q(\mathbf{x})}{p'(\mathbf{x})}\nabla_\theta\log p_\theta (\mathbf{x})]\\
&=\frac{1}{Z}\mathbb{E}_{p_\theta}[r_D(\mathbf{x})\nabla_\theta\log p_\theta (\mathbf{x})] 
\end{align*}
where we assume that $p'=p_\theta$ and the delayed generator is only one step behind the current update in the experiments. This importance sampling procedure was discovered independently from us
by~\citep{hjelm2017boundary}. We propose to optimize the generator using the following novel gradient estimator: 
\begin{align}
\small
\nabla L_G(\theta) &\approx  \sum_{i=1}^m (\frac{r_D(\mathbf{x}_i)}{\sum_i r_D(\mathbf{x}_i)}-b)\nabla \log p_\theta(\mathbf{x}_i) = E(\{\mathbf{x}_i\}_1^m)
\label{eq:2}
\end{align}
where $b$ is a baseline from reinforcement learning in order to reduce variance. In practice, we let $b$ increase very slowly from 0 to 1. Combined with the objective of the discriminator in an ordinary GAN, we get the proposed MaliGAN algorithm as shown in Algorithm~\ref{algo-1}. 

\vspace{0.2cm}
\begin{algorithm}
\caption{MaliGAN}
{\fontsize{9}{9}\selectfont
\begin{algorithmic}[1]
\REQUIRE  \begin{varwidth}[t]{\linewidth}
A generator $p$ with parameters $\theta$. \par
A discriminator $D(x)$ with parameters $\theta_d$.\par
A baseline $b$.
\end{varwidth} 

\FOR{number of training iterations}
    \FOR{k steps}
    \STATE Sample a minibatch of samples $\{\mathbf{x}_i\}_{i=1}^m$ from $p_\theta$.
    \STATE Sample a minibatch of samples $\{\mathbf{y}_i\}_{i=1}^m$ from $p_d$.
    \STATE Update the parameter of discriminator by taking gradient ascend of discriminator loss 
    \small
    \begin{equation*}
    \sum_i [\nabla_{\theta_d}\log D(\mathbf{y}_i)] + \sum_i[\nabla_{\theta_d}\log (1-D(\mathbf{x_i}))]
    \end{equation*}
    \normalsize
    \ENDFOR
    \STATE Sample a minibatch of samples $\{\mathbf{x}_i\}_{i=1}^m$ from $p_\theta$.
    \STATE Update the generator by applying gradient update 
    \small
    \begin{equation*}
 \sum_{i=1}^m (\frac{r_D(\mathbf{x}_i)}{\sum_i r_D(\mathbf{x}_i)}-b)\nabla \log p_\theta(\mathbf{x}_i)
    \end{equation*}
    \normalsize
\ENDFOR
\end{algorithmic}
}
\label{algo-1}
\end{algorithm}

\subsection{Analysis}
\label{sec:analysis}
The proposed objective in Eq.~\ref{eq:2} is also theoretically guaranteed to be sound. In the following theorem, we show that our training objective approximately optimizes the KL divergence $\text{KL}(q(\mathbf{x})||p_\theta(\mathbf{x}))$ when $D$ is close to optimal. What's more, the objective still makes sense when $D$ is well trained but far from optimal. 
\begin{theorem}
\label{eq:8}
We have the following two theoretical guarantees for our new training objective. 
\item[(i)]
If discriminator $D(\mathbf{x})$ is optimal between delayed generator $p'$ and real data distribution $p_d$, we have the following equation. 
\begin{align*}
\small
\mathbb{E}_{p_d}[ \log p_\theta(\mathbf{x})]=\frac{1}{Z(\theta')}\mathbb{E}_{p'}[r_D(\mathbf{x})\log p_\theta(\mathbf{x})]
\end{align*}
where $Z(\theta')=\mathbb{E}_{p'}[r_D(\mathbf{x})]=1$.

\item[(ii)] If $D(x)$ is trained well but not sufficiently, namely, $\forall x$, $D(x)$ lies between 0.5 and  $\frac{p_d}{p_d+p'}$, we have the property that for $m \rightarrow \infty$, almost surely
\begin{equation}
E(\{\mathbf{x_i}\}_1^m)\cdot \nabla_\theta \text{KL}(p_d||p_\theta) > 0
\end{equation}
The above gives us a condition for our objective to still push the generator in a descent direction even when the discriminator is not trained to optimality. 
\end{theorem}

In addition to its attractiveness in theory, we now demonstrate why the gradient estimator in Eq.~\ref{eq:2} of $\nabla L_G(\theta)$ practically can produce better training signal for the generator than the original GAN objective. Similar discussions can be found in~\cite{Bornschein2014ReweightedW,Norouzi2016RewardAM}. 

In the original GAN setting from a reinforcement learning perspective, e.g. the inclusive KL in Eq.~\ref{eq:3}, the free running auto-regressive model can be viewed as an RL agent exploring the state space and getting a reward, $D$ or $\log D$, at the end of the exploration. The model then tries to adjust the probability of each of its exploration paths according to this reward. However, this gradient estimator would be drastically inefficient when almost all generated paths had a very small discriminator output. Unfortunately, this is very common in GAN training and cannot even be solved with a carefully selected baseline. 

In the MaliGAN objective, however, the partition function $Z$ is estimated using the samples from the minibatch, which helps dealing with the above dilemma. When we choose, for example, baseline $b=1$, we can see that the sum of the weights on the generated paths are zero, and the probability of each path is adjusted not according to the absolute value of the discriminator output, but its relative quality in that minibatch. This ensures that the model can always learn something as long as there exist some generations better than others in that mini-batch. Furthermore, the previous theorem ensures the consistency of the mini-batch level normalization procedure.

From a theoretical point of view, this normalization procedure also helps. Although at the first glance, when $D$ is optimal, one can prove that $Z=1$, so estimating $Z$ seems to only introduce additional variance to the model. However, using this estimator in fact reduces the variance due to the following reason: $r_D(\mathbf{x})$ is actually a function with singularity when $\mathbf{x}$ is in a region $\Omega$ in the data space on which $D(\mathbf{x})\approx 1$. Even with very careful pre-training, such a region $\Omega$ $r_D \gg 0$ and $p'(\Omega)\approx 0$, making the ratio blow up. In our target $\frac{1}{Z(\theta')}\mathbb{E}_{p'}[r_D(\mathbf{x})\log p_\theta(\mathbf{x})]$, since it is almost impossible to get samples from $\Omega$ with $p'$ in a reasonable size mini-batch, the actual distribution we are sampling from is a ``regularized'' distribution $p_{\backslash\Omega}$ where $p_{\backslash\Omega}(\Omega)=0$ and $p_{\backslash\Omega}\approx p'$. So when doing importance sampling to estimate our training objective $\nabla L_G = \mathbb{E}_{p_d}[\nabla_\theta\log p_\theta (\mathbf{x})]$ with small mini-batches, we are actually doing normalized-weights importance sampling based on $p_{\backslash\Omega}$: $\nabla L_G \approx \mathbb{E}_{p_{\backslash\Omega}}[r_D(\mathbf{x})\nabla_\theta\log p_\theta (\mathbf{x})]/\mathbb{E}_{p_{\backslash\Omega}}[r_D(\mathbf{x})]$. Since the Monte Carlo estimator has much more variance to estimate $\mathbb{E}_{p'}[r_D(\mathbf{x})\nabla_\theta\log p_\theta (\mathbf{x})]$ than 
$\mathbb{E}_{p_{\backslash\Omega}}[r_D(\mathbf{x})\nabla_\theta\log p_\theta (\mathbf{x})]$, in practical mini-batch training settings, we can view that we are doing importance sampling with the distribution $p_{\backslash\Omega}$, and this objective has much less variance compared to importance sampling with $p'$ on $r_D$ which has an infinite singularity. This is why estimating $Z= \mathbb{E}_{p_{\backslash\Omega}}[r_D(\mathbf{x})]$ is important in order to reduce the variance in the mini-batch training setting. 

When training auto-regressive models with teacher forcing, a serious problem is exposure bias \citep{ranzato2015sequence,rewardaug,lamb2016professor}. Namely, the model is only trained on demonstrated behaviors (real data samples), but we also want it to be trained on free-running behaviors. When we set a positive baseline $b>0$, the model first generates $m$ samples, and then tries to adjust the probabilities of each generated samples by trying to reinforce the best behaviors and exclude the worse behaviors relatively to those in the mini-batch. 

\subsection{Variance Reduction in MaliGAN}
\label{sec:tricks}
The proposed renormalized objective in MaliGAN supports much more stable training behavior than the RL objective in a standard GAN. Nevertheless, when the long sequence generation procedure consists of multiple steps of random sampling, we find it is better to further integrate the following advanced variance reduction techniques.

\subsubsection{Monte Carlo Tree Search}
Instead of using the same weight for all time steps in one sample, we use the following formula which is well known in the RL literature:
\begin{align*}
\small
\mathbb{E}_{p_\theta}[r_D(\mathbf{x})\nabla p(\mathbf{x})]=\mathbb{E}_{p_\theta}[\sum_{t=1}^L Q(a_t,\mathbf{s}_t)\nabla p_\theta(a_t|\mathbf{s}_t)]
\end{align*}
where $Q(a,\mathbf{s})$ stands for the ``expected total reward'' given by $r_D=\frac{D}{1-D}$ of generating token $a$ given previous generation $\mathbf{s}$, which can be estimated with, e.g., Monte Carlo tree search (MCTS, ~\citet{alphago}).

Thus, following the gradient estimator presented in Theorem~\ref{eq:8}, we derive another gradient estimator:
\begin{align*}
\small
\nabla L_G(\theta) \approx \frac{\sum_i L_i}{m\sum Q(a^i_{t},\mathbf{s}^i_t)} \sum_{i,t}^{m,L_i} Q(a^i_{t},\mathbf{s}^i_t)\nabla \log p_\theta(a^i_t|\mathbf{s}^i_t)
\end{align*}
where $m$ is the size of the mini-batch. Using Monte Carlo tree search brings in several benefits. First, it allows different steps of the generated sample to be adjusted with different weights. Second, it gives us a more stable estimator of the partition function $Z$. Both of these two properties can dramatically reduce the variance of our proposed estimator. 

\subsubsection{Mixed MLE-Mali Training}
When dealing with long sequences, the above model may result in accumulated variance. To alleviate the issue, we significantly reduce the variance by clamping the input using the training data for $N$ time steps, and switch to free running mode for the remaining $T-N$ time steps. Then during our training procedure, inspired from~\citet{ranzato2015sequence}, we slowly move $N$ from $T$ towards 0.

The training objective is equivalent to setting $q$ in the last section to:
\begin{align*}
\small
q(x_0,x_1,\cdots x_L) = p_d(x_0,\cdots x_N) q(x_{N+1},\cdots x_L|x_0,\cdots x_N) 
\end{align*}

We also assume $D$ is trained on the real samples and fake samples generated by 
\begin{align*}
\small
p_f(x_0,\cdots x_L) = p_d(x_0,\cdots x_N) p_\theta(x_{N+1},\cdots x_L|x_0,\cdots x_N) 
\end{align*}

Let $\mathbf{x}_{\leq N}=(x_0,x_1,\cdots x_N),\mathbf{x}_{>N}=(x_{N+1},\cdots x_L)$, we have:
\begin{align*}
\small
\nabla L_{G}=&\mathbb{E}_q[\nabla \log p_\theta(\mathbf{x})]\\ 
=&\mathbb{E}_{p_d}[\nabla \log p_\theta(\mathbf{x}_{\leq N})] + \mathbb{E}_q[\nabla\log p_\theta (\mathbf{x}_{>N}|\mathbf{x}_{<N})] \\
=&\mathbb{E}_{p_d}[\nabla \log p_\theta(x_0,x_1,\cdots x_T)]\\
&+\frac{1}{Z}\mathbb{E}_{p_\theta}[ \sum_{t=N+1}^{L} r_D(\mathbf{x})\nabla \log p_\theta(a_t|\mathbf{s}_t)]
\end{align*}

For each sample $\mathbf{x}_i$ from the real data batch, if it has length larger than $N$, we fix the first $N$ words of $\mathbf{x}_i$, and then sample $n$ times from our model till the end of the sequence, and get $n$ samples $\{\mathbf{x}_{i,j}\}_{j=1}^n$.

We then have the following series of mini-batch estimators for each $0\leq N\leq T$:
\begin{align}
\label{eq:4}
\begin{split}
\small
\nabla L^N_{G} \approx & \sum_{i=1,j=1}^{m,n} (\frac{r_D(\mathbf{x}_{i,j})}{\sum_j r_D(\mathbf{x}_{i,j})}-b)  \nabla\log p_\theta(\mathbf{x}^{>N}_{i,j}|\mathbf{x}^{\leq N}_{i}) \\
&+\frac{1}{m}\sum_{i=1}^m\sum_{t=0}^N p_\theta(a^i_t|\mathbf{s}^i_t) =  E_N(\mathbf{x}_{i,j}) 
\end{split}
\end{align}

One difference is that in this model, we normalize the coefficients $r_D(\mathbf{x}_{i,j})$ based only on samples generated from a single real data sample $\mathbf{x}_{i}$. The reason of using this trick will be explained in next sub-section.  

We have the following theorem which guarantees the theoretical property of this estimator. 

\begin{theorem}
When $D$ is correctly trained but not optimal in the sense of Theorem ~\ref{eq:8}, when $m\rightarrow \infty$, we almost surely have $\forall 0\leq N\leq T$,
\begin{equation}
E_N(\mathbf{x}_{i,j})\cdot \nabla_\theta \text{KL}(p_d||p_\theta) >0
\end{equation}
\end{theorem}

\subsubsection{Single real data based renormalization}
Many generative models have multiple layers of randomnesses. For example, in auto-regressive models, the samples are generated via multiple sampling steps. Other examples include hierarchical generative models like deep Boltzmann machines and deep belief networks~\cite{dbm,dbf}.

In these models, high-level random variables are usually responsible for modeling high-level decisions or ``modes'' of the probability distribution. Changing them can result in much larger effects than that from changing low-level variables. Motivated by this observation, in each mini-batch we first draw a mini-batch of samples (e.g. 32) of high-level latent variables, and then for each high level value we draw a number of low level data samples (e.g. 32). Then we re-estimate the partition function $Z$ from the low-level samples that are generated by each high-level samples. Because lower-level sampling has a much smaller variance, the model can receive better gradient signals from the weights provided by the discriminator.

This sampling principle is corresponding to applying the mixed MLE-Mali training discussed above in the auto-regressive settings. In this case we first sample a few data samples, then fix the first $N$ words and let the network generate a lot of samples after $N$ as our next mini-batch. We refer this full algorithm to sequential MaliGAN with Mixed MLE Training, which is summarized in Algorithm~\ref{algo-2}.

\vspace{0.2cm}
\begin{algorithm}
\caption{Sequential MaliGAN with Mixed MLE Training}
{\fontsize{9}{9}\selectfont
\begin{algorithmic}[1]
\REQUIRE  \begin{varwidth}[t]{\linewidth}
A generator $p$ with parameters $\theta$. \par
A discriminator $D(x)$ with parameters $\theta_d$.\par
Maximum sequence length $T$, step size $K$. \par
A baseline $b$, sampling multiplicity $m$.
\end{varwidth} 
\STATE $N=T$
\STATE Optional: Pretrain model using pure MLE with some epochs.
\FOR{number of training iterations}
    \STATE $N$ = $N$ - $K$
    \FOR{k steps}
    \STATE Sample a minibatch of sequences $\{\mathbf{y}_i\}_{i=1}^m$ from $p_d$.
    \STATE While keeping the first $N$ steps the same as $\{\mathbf{y}_i\}_{i=1}^m$, sample a minibatch of sequences $\{\mathbf{x}_i\}_{i=1}^m$ from $p_\theta$ from time step $N$.
    \STATE Update the discriminator by taking gradient ascend of discriminator loss.
    \small
    \begin{equation*}
    \sum_i [\nabla_{\theta_d}\log D(\mathbf{y}_i)] + \sum_i[\nabla_{\theta_d}\log (1-D(\mathbf{x_i}))]
    \end{equation*}
    \normalsize
    \ENDFOR
    \STATE Sample a minibatch of sequences $\{\mathbf{x}_i\}_{i=1}^m$ from $p_d$.
    \STATE For each sample $\mathbf{x}_i$ with length larger than $N$ in the mini-batch, clamp the generator to the first $N$ words of $s$, and freely run the model to generate $m$ samples $\mathbf{x}_{i,j},j=1,\cdots m$ till the end of the sequence.
    \STATE Update the generator by applying the mixed MLE-Mali gradient update 
    \small
    \begin{equation*}
    \begin{aligned}
\nabla L^N_{G} \approx & \sum_{i=1,j=1}^{m,n} (\frac{r_D(\mathbf{x}_{i,j})}{\sum_j r_D(\mathbf{x}_{i,j})}-b)  \nabla\log p_\theta(\mathbf{x}^{>N}_{i,j}|\mathbf{x}^{\leq N}_{i}) \\
&+\frac{1}{m}\sum_{i=1}^m\sum_{t=0}^N p_\theta(a^i_t|\mathbf{s}^i_t) 
	\end{aligned}
    \end{equation*}
    \normalsize
\ENDFOR
\end{algorithmic}
}
\label{algo-2}
\end{algorithm}

The reason why doing this single real sample based renormalization is beneficial can be summarized around two elements. First, consider $S$ is a sample from the training set. The first N words $S_{\leq N}$ should be completed by our model. The conditional distribution $p_d(S'_{>N}|S_{\leq N})$ should be much simpler than the full distribution $p_d$. Namely, $p_d(S'_{>N}|S_{\leq N})$ consists of only one or a few ``modes''. So this renormalization technique can be viewed as trying to train the model on these simpler conditional distributions, which gives more stable gradients. 

Second, this normalization scheme makes our model robust to mode missing, which is a common failure pattern when training GANs~\citep{Che2016ModeRG}. Single sample based renormalization ensures that for every real sample $S$, the model can receive a moderately strong training signal for how to perform better on generating $S_{>N}$ conditioned on $S_{\leq N}$. However, in batch-wise renormalization as in the basic MaliGAN, this is not possible because there might be some completions $S'$ with $r_D(S')$ very large, so other training samples in that mini-batch receives very little gradient signals.

\section{Experiments}
\label{sec:exp}
To examine the effectiveness of the proposed algorithms, we conduct experiments on three discrete sequence generation tasks. We achieve promising results on all three tasks, including a standard and challenging language modeling task. From the empirical results and the following analysis, we demonstrate the soundness of MaliGAN and show its robustness to overfitting. 

\subsection{Discrete MNIST}
We first evaluate MaliGAN on the binarized image generation task for the MNIST hand-written digits dataset, similar with~\citet{hjelm2017boundary}. The original datasets have 60,000 and 10,000 samples in the training and testing sets, respectively. We split the training set and randomly selected 10,000 samples for validation. We adopted as the generator a deep convolutional neural network based on the DCGAN architecture~\cite{DCGAN}. To generate the discrete samples, we sample from the generator's output binomial distribution. We adopt Algorithm~\ref{algo-1} of MaliGAN for training and use the single latent variable renormalization technique for variance reduction.

\begin{figure}[h]
    \centering
    \begin{minipage}{0.24\textwidth}
        \centering
        \includegraphics[width=0.9\textwidth]{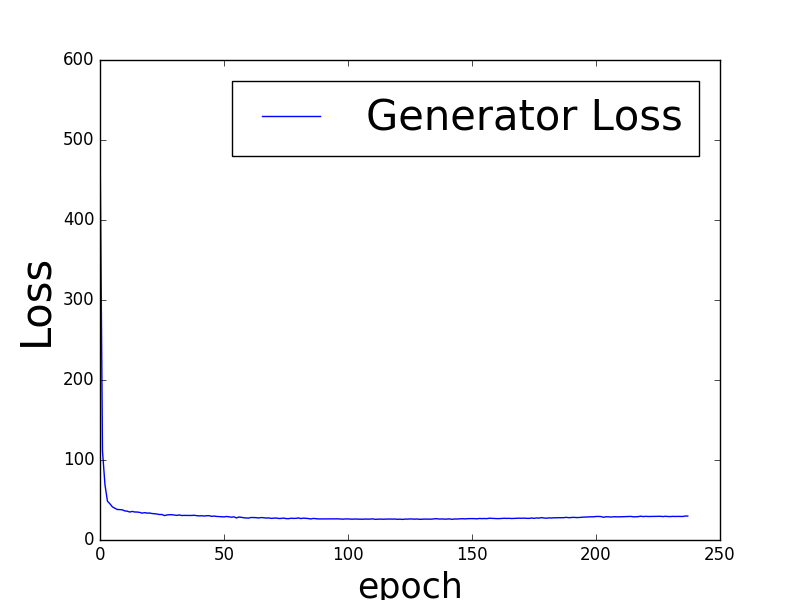}
    \end{minipage}\hfill
    \begin{minipage}{0.24\textwidth}
        \centering
        \includegraphics[width=0.9\textwidth]{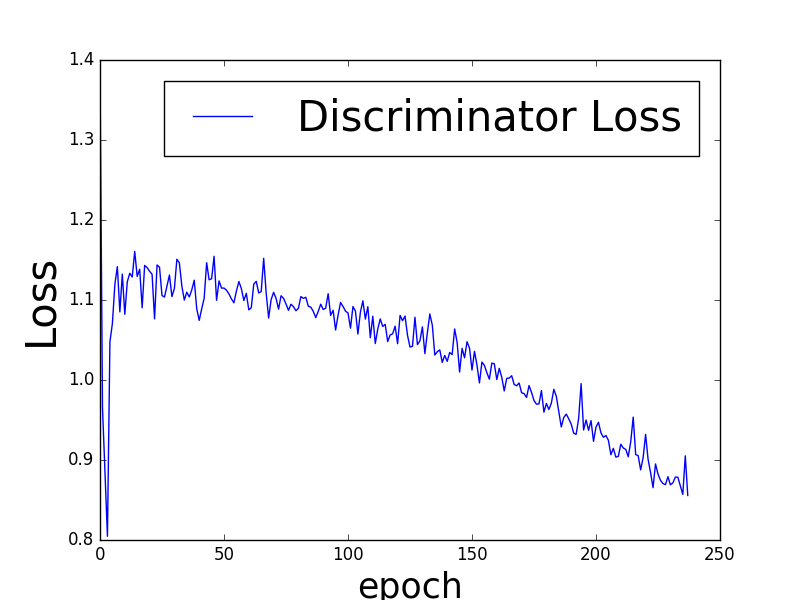}
    \end{minipage}
     \caption{The training loss of the generator (left) and the discriminator (right) of MaliGAN on Discrete MNIST task.}
    \label{fig:maligan-mnist-loss}
\end{figure}

\begin{figure}[h]
    \centering
    \begin{minipage}{0.24\textwidth}
        \centering
        \includegraphics[width=0.75\textwidth]{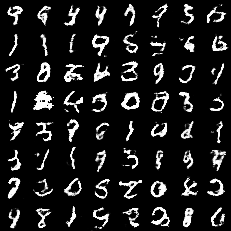}
    \end{minipage}\hfill
    \begin{minipage}{0.24\textwidth}
        \centering
        \includegraphics[width=0.75\textwidth]{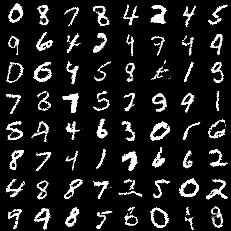}
    \end{minipage}
     \caption{Samples generated by REINFORCE-like model (left) and by MaliGAN (right).}
     \label{fig:samples_mnist}
\end{figure}

To compare our proposed MaliGAN with the models trained using the discriminator's output as a direct reward, we also train a generator with the same network architecture, but use the output of the discriminator as the weight of generated samples. We denote it as the REINFORCE-like model. The comparison results are shown in Figure~\ref{fig:maligan-mnist-loss} and Figure~\ref{fig:samples_mnist}.

The two figures in the first line are training losses of the generator and discriminator from the proposed MaliGAN. We can see the training process of MaliGAN with variance reduction techniques is stable and the loss curve is meaningful. The bottom two figures in Figure~\ref{fig:samples_mnist} are samples generated by the REINFORCE-like model and by MaliGAN. Clearly, the samples generated by MaliGAN have much better visual quality and resemble closely the training data.

\vspace{0.4cm}

\subsection{Poem Generation}
We examine the effectiveness of our model on a Chinese poem generation task. Typically, there are two genres of Chinese poems. We refer with \textit{Poem-5} and \textit{Poem-7} to those consisting of 5 or 7 Chinese characters each in a short sentence, respectively. We use the dataset provided in~\cite{poetry}, and split them in the standard way~\footnote{\url{http://homepages.inf.ed.ac.uk/mlap/Data/EMNLP14/}}. 

The generator is a one-layer LSTM~\cite{lstm} with 32 hidden units for Poem-5 and 100 for Poem-7. Our discriminators are two-layer Bi-LSTMs with 32 hidden neurons. We denote our models trained with Algorithm~\ref{algo-1} and Algorithm~\ref{algo-2} as \textbf{MaliGAN-basic} and \textbf{MaliGAN-full}. We choose two compared models, the auto-regressive model with same architecture but trained with maximum likelihood (MLE), and SeqGAN~\cite{seqgan}. Following~\citet{seqgan}, we report the BLEU-2 scores in Table~\ref{poem_result}~\cite{bleu}.  

\begin{figure}[h]
    \centering
    \begin{minipage}{0.24\textwidth}
        \centering
        \includegraphics[width=0.9\textwidth]{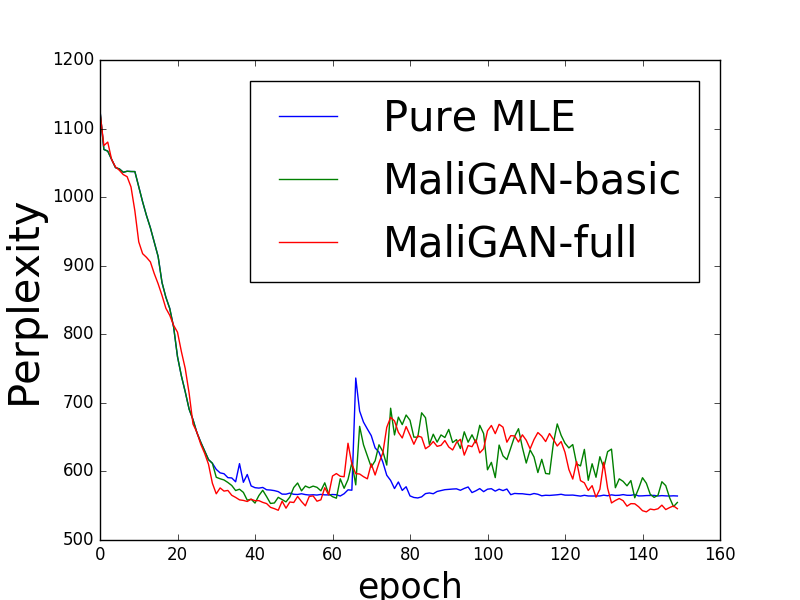}
    \end{minipage}\hfill
    \begin{minipage}{0.24\textwidth}
        \centering
        \includegraphics[width=0.9\textwidth]{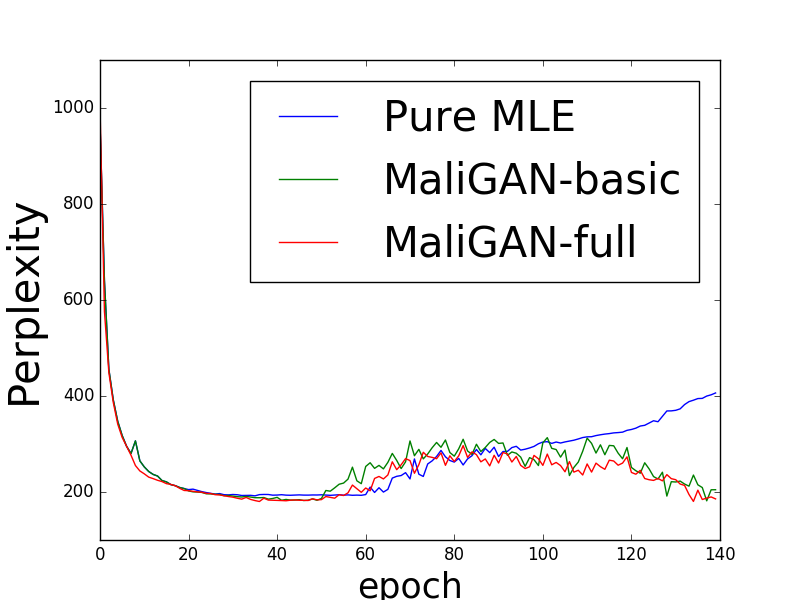}
    \end{minipage}
      \caption{Perplexity curves on Poem-5 (left) and Poem-7 (right).}
      \label{fig:ppl_7}
\end{figure}

MaliGAN-full obtained the best BLEU-2 scores on par on both tasks, and MaliGAN-basic was the next best. Clearly, MLE lagged far behind despite the same architecture, which should be attributed to the inherent defect in the MLE teacher-forcing training framework. As pointed by previous researchers~\citet{wiseman2016sequence}, BLEU might not be a proper evaluation metric, we also calculate the Perplexity of these four models, obtaining qualitatively similar results. The best scores are reported in Table~\ref{poem_result} and the Perplexity curves are illustrated in Figure~\ref{fig:ppl_7}.

\begin{table}[h]
\small
\label{poem_result}
\caption{Experimental results on Poetry Generation task. The result of SeqGAN is directly taken from~\citep{seqgan}.}
\vspace{0.3cm}
\centering
\begin{tabular}{rcccc}
\toprule
\centering{\multirow{2}{*}{Model}} & \multicolumn{2}{c}{Poem-5}     & \multicolumn{2}{c}{Poem-7} \\
\cmidrule{2-5}
& BLEU-2 & PPL & BLEU-2 & PPL\\
\midrule
MLE & 0.6934 & 564.1 & 0.3186 & 192.7 \\
{SeqGAN} & 0.7389 & - & - & -\\
{MaliGAN-basic} & 0.7406 & 548.6& 0.4892 & 182.2\\
\bf{MaliGAN-full} & \bf{0.7628} & \bf{542.7}& \bf{0.5526} &\bf{180.2} \\\hline
\end{tabular}
\end{table}

From the above figures, we can see how our models perform during the training procedure. Although with some oscillations, both MaliGAN-basic and MaliGAN-full achieved lower perplexity. Especially on Poem-7 from Figure~\ref{fig:ppl_7}, our proposed models both prevent overfitting when MLE ended up with that. A comparison between the training curve of MaliGAN-basic and that of MaliGAN-full, we can find that the latter has less variance. This demonstrates the effectiveness of the advanced variance reduction techniques in our full model. The peak in the MLE curve on Poem-5 in Figure~\ref{fig:ppl_7} is, however, unlikely to be a result of overfitting because that MLE ``recovered'' from it fast and continued to convergence till the end. In fact, we find it harder to train a stable MLE model on Poem-5 than on Poem-7. We conjecture this resulted from the intricate mutual influence between the improper evaluation and the small training data size.

\subsection{Sentence-Level Language Modeling}
We also examine the proposed algorithm on a more challenging task, sentence-level language modeling, which can be considered as a fundamental task with applications to various discrete sequence generation tasks. To explore the possibilities and limitations of our algorithm, we conduct extensive experiments on the standard Penn Treebank (PTB) dataset~\cite{ptb} through parameter searching and model ablations. For evaluation we report sentence-level perplexity, which is the averaged perplexity on all sentences in the test set. For simplicity and efficiency, we adopt a 1-layer GRU~\cite{gru} as our generator, and set the same setting for the baseline model trained with standard teacher forcing\cite{teacherforcing}. We use a Bi-directional GRU network as our discriminator. To stabilize training and provide good initialization for the generator, we first pre-train our generator on the training set using teacher forcing, then we train two models, MaliGAN-basic and MaliGAN-full. MaliGAN-basic is trained with Algorithm~\ref{algo-1} without MCTS. MaliGAN-full is trained by Algorithm~\ref{algo-2} with all the variance reduction techniques included.

Note that the computational cost of MCTS is very large, so we remove all sentences longer than 35 words in the training set. We set $N = 30$ and $K = 5$ at the beginning of the training and pre-train our discriminator to make it reliable enough to provide informative and correct signals for the generator. The perplexity shown in Table~\ref{lm_result} is achieved by our best performing model, which has 200 hidden neurons and 200 dimensions for word embeddings.

\begin{table}[h]
\small
\label{lm_result}
\caption{Experimental results on PTB. Note that we evaluate the models in sentence-level.}
\vspace{0.3cm}
\centering
\begin{tabular}{rccc}
\toprule
& MLE & MaliGAN-basic & \bf{MaliGAN-full} \\
\midrule
Valid-Perplexity & 141.9 &  131.6 & \bf{128.0} \\
Test-Perplexity & 138.2 &  125.3 & \bf{123.8} \\\hline
\end{tabular}
\end{table}

From Table~\ref{lm_result} we can see, the simplest model trained by MaliGAN reduced the perplexity of the baseline effectively. Both the basic and the full model, i.e., MaliGAN-basic and MaliGAN-full obtained a notably lower perplexity compared with the MLE model. Although the PTB dataset is much more difficult, we obtain results consistent with Table~\ref{poem_result}. It is encouraging to see that our model is more robust to overfitting in consideration of the relative small size of the PTB data. These results strengthen our belief to realize our algorithm on even larger datasets, which we leave as a future work.

The positive result again demonstrates the effectiveness of MaliGAN, whose primary component is the novel optimization objective we propose in Eq.~\ref{eq:2}. Besides, we also gain insights from the model ablation tests about the advanced variance reduction techniques provided in Section~\ref{sec:tricks}. Combined with the Perplexity curve in Figure~\ref{fig:ppl_7}, we can see that with advanced techniques, MaliGAN-full performed in a more stable way during training and can to some extent achieve lower perplexity scores than MaliGAN-basic. We believe these fruitful techniques will be beneficial in other similar problem settings. 

\section{Related Work}
To improve the performance of discrete auto-regressive models, some researchers aim to tackle the \textit{exposure bias} problem, which is discussed detailed in~\cite{ranzato2015sequence,serban2016hred,wiseman2016sequence}. The problem occurs when the training algorithm prohibits models to be exposed to their own predictions during training. The second issue is the discrepancy between the objective during training and the evaluation metric during testing, which is analyzed in~\citet{ranzato2015sequence} and then summarized as \textit{Loss-Evaluation Mismatch} by~\citet{wiseman2016sequence}. Typically, the objectives in training auto-regressive models are to maximize the word-level probabilities, while in test-time, we often evaluate the models using sequence-level metrics, such as BLEU~\cite{bleu}. To alleviate these two issues, the most straightforward way is to add the evaluation metrics into the objective in the training phase. Because these metrics are often discrete which cannot be utilized through standard back-propagation, researchers generally seek help from reinforcement learning. \citet{ranzato2015sequence} exploits REINFORCE algorithm~\cite{reinforce} and proposes several model variants to well situate the algorithm in text generation applications. \citet{rlcaption} shares similar idea and directly optimizes image caption metrics through policy gradient methods~\cite{pg}. There exists a third issue, namely \textit{Label Bias}, especially in sequence-to-sequence learning framework, which obstacles the MLE trained models to be optimized globally~\cite{Andor2016GloballyNT,wiseman2016sequence} 

To addresses the abovementioned issues in training auto-regressive models, we propose to formulate the problem under the setting of generative adversarial networks. Initially proposed by~\citet{gan}, generative adversarial network (GAN) has attracted a lot of attention because it provides a powerful framework to generate promising samples through a min-max game. Researchers have successfully applied GAN to generate promising images conditionally~\cite{CGAN,reed2016generative,stackgan} and unconditionally~\cite{DCGAN,nguyen2016ppgn}, to realize image manipulation and super-resolution~\cite{zhu2016generative,sonderby2016amortised,photoresolution}, and to produce video sequences~\cite{ganvideo,zhou2016learning,saito2016temporal}. Despite these successes, the feasibility and advantage on applying GAN to text generation are restrictedly explored yet noteworthy. 

It is appealing to generate discrete sequences using GAN as discussed above. The generative models are able to utilize the discriminator's output to make up the information of its own distribution, which is inaccessible if trained by teacher forcing~\cite{teacherforcing,ranzato2015sequence}. However, it is nontrivial to train GAN on discrete data due to its discontinuity nature. The instability inherent in GAN training makes things even worse~\cite{improved,Che2016ModeRG,towards,WGAN}. \citet{lamb2016professor} exploits adversarial domain adaption to regularize the training of recurrent neural networks. \citet{seqgan} applies GAN to discrete sequence generation by directly optimizing the discrete discriminator's rewards. They adopt Monte Carlo tree search technique~\cite{alphago}. Similar technique has been employed in~\citet{li2017adversarial} which improves response generation by using adversarial learning. 

In~\citet{Bornschein2014ReweightedW}, which inspired us, the authors propose a way of doing mini-batch reweighting when training latent variable models with discrete variables. However, they make use of inference network which are infeasible in the GAN setting.

Our work is also closely related to~\citet{rewardaug}. In~\citet{rewardaug}, they propose to work with the objective $\text{KL}(p_d||p_\theta)$ in a conditional generation setting. In this case, the situation is similar with ours because rewards such as BLEU scores are available. However, conditional generation metrics such as BLEU scores are decomposable to each time steps, so this property can make them able to directly sample from the augmented distributions, which is not possible for sequence-level GANs, e.g., language modeling. So we have to use importance sampling to train the model. 

\section{Discussions and Future Work}
In spite of their great popularity on continuous datasets such as images, GANs haven't yet achieved an equivalent success in discrete domains such as natural language processing. We observed that the main cause of this gap is that while the discriminator can almost perfectly discriminate the good samples from the bad ones, it is notoriously difficult to pass this information to the generator due to the difficulty of credit assignment through discrete computation and inherent instability of RL algorithms applied to dynamic environments with sparse reward. 

In this work, we take a different approach. We start first from the maximum likelihood training objective $\text{KL}(p_d||p_\theta)$, and then use importance sampling combined with the discriminator output to derive a novel training objective. We argue that although this objective looks similar to the objective used in reinforcement learning, the normalization in fact does reduce the variance of the estimator by ignoring the region $\Omega$ in the data space around the singularity of $r_D$ in which the generator $p_\theta$ has almost zero probability to get samples from. Namely, by estimating the partition function $Z$ using samples, we are approximately doing normalized importance sampling with another distribution $p_{\backslash\Omega}$ which has much lower variance c.f. Section~\ref{sec:analysis}. Practically, this single real sample normalization process combined with mixed training~\citep{ranzato2015sequence} successfully avoided the missing mode problem by providing equivalent training signal for each mode. 

Besides successfully reducing the variances of normal reinforcement learning algorithms, our algorithm is surprisingly robust to overfitting. Teacher forcing is prone to overfit, because by maximizing the likelihood of the training data, the model can easily fit not only the regularities but also the noise in the data. However in our model, if the generator tries to fit too much noise in the data, the generated sample will not look good and hopefully the discriminator will be able to capture the differences between the generated and the real samples very easily.

As for future work, we are going to train the model on large datasets such as Google's one billion words~\cite{billion} and on conditional generation cases such as dialogue generation. 

\bibliography{maligan}
\bibliographystyle{icml2017}

\end{document}